\documentclass[conference]{IEEEtran}
\IEEEoverridecommandlockouts
\usepackage{float}
\usepackage{hyperref}

\usepackage{cite}
\usepackage{amsmath,amssymb,amsfonts}
\usepackage{algorithmic}
\usepackage{graphicx}
\usepackage{textcomp}
\usepackage{xcolor}
\def\BibTeX{{\rm B\kern-.05em{\sc i\kern-.025em b}\kern-.08em
    T\kern-.1667em\lower.7ex\hbox{E}\kern-.125emX}}
    
\usepackage{fancyhdr}
\begin{document}

\title{Fine-Grained Emotion Prediction by Modeling Emotion Definitions}

\author{\IEEEauthorblockN{Gargi Singh\IEEEauthorrefmark{1},
Dhanajit Brahma\IEEEauthorrefmark{2}, 
Piyush Rai\IEEEauthorrefmark{3}, 
Ashutosh Modi\IEEEauthorrefmark{4}}
\IEEEauthorblockA{CSE Department,
Indian Institute of Technology Kanpur (IIT-K), 
Kanpur 208016, India \\
Email: \IEEEauthorrefmark{1}sgargi@iitk.ac.in,
\IEEEauthorrefmark{2}dhanajit@cse.iitk.ac.in,
\IEEEauthorrefmark{3}piyush@cse.iitk.ac.in,
\IEEEauthorrefmark{4}ashutoshm@cse.iitk.ac.in}}

\maketitle
\thispagestyle{fancy}

\begin{abstract}
In this paper, we propose a new framework for fine-grained emotion prediction in the text through emotion definition modeling. Our approach involves a multi-task learning framework that models definitions of emotions as an auxiliary task while being trained on the primary task of emotion prediction. We model definitions using masked language modeling and class definition prediction tasks. Our models outperform existing state-of-the-art for fine-grained emotion dataset GoEmotions.  We further show that this trained model can be used for transfer learning on other benchmark datasets in emotion prediction with varying emotion label sets, domains, and sizes. The proposed models outperform the baselines on transfer learning experiments demonstrating the generalization capability of the models. 
\end{abstract}


\section{Introduction}
    %Field introduction
Recent advancements in Artificial Intelligence (AI) have made machines/computers an integral part of our lives, e.g., personal digital assistants. These AI based technologies are getting better at understanding what humans express explicitly via language (both speech and text). However, humans use language not just to convey information but also to express their inner feelings and mental states. The majority of the time, humans express emotions in language implicitly. AI pioneer Marvin Minsky in his book ``The Emotion Machine" \cite{minsky2007emotion}, has explained the importance of emotions and how it is not that different from the underlying process of thinking in humans. To develop truly intelligent systems that understand and assist humans, we would like these machines to recognize and understand implicit emotions in the interactions. Humans do emotion recognition effortlessly; however, for machines, it is not a trivial activity. Intending to develop emotionally intelligent machines, the fundamental problems of emotion recognition \cite{b1, b2, witon2018disney, singh2021end} and generation \cite{goswamy2020adapting} are active areas of research in the AI community. 

    \begin{table}[]
        \begin{center}
        \begin{tabular}{ll}
        \hline
        \textbf{Text} & \textbf{Label(s)} \\ \hline
        \begin{tabular}[c]{@{}l@{}}It's better to say a moment like that could truly \\ ignite her love for the game rather than \\ putting a bit of a damper on it.\end{tabular}                    & love                                                             \\ \hline
        \begin{tabular}[c]{@{}l@{}}I did hear that {[}NAME{]} is friends with {[}NAME{]},\\  I think on the SC sub. This was probably like\\  a month ago. Could be.\end{tabular}                    & confusion                                                        \\ \hline
        \begin{tabular}[c]{@{}l@{}}Sometimes life actually hands you lemons. \\ We're just lucky that we have a proverbial phrase\\  that gives us an idea of what we can do with them.\end{tabular} & \begin{tabular}[c]{@{}l@{}}approval, \\ realization\end{tabular} \\ \hline
        \begin{tabular}[c]{@{}l@{}}Does nobody notice that this is a doctored photo..??\\  Like really bad clone stamp.. I mean it could be \\ beautiful but this is not a real photo.\end{tabular}  & disapproval                                                      \\ \hline
        \begin{tabular}[c]{@{}l@{}}I think the fan base is mostly past that at this point.\\  Almost everyone has MASSIVE problems with \\ some of the decisions Nintendo makes.\end{tabular}        & neutral                                                          \\ \hline
        \end{tabular}
        \vspace*{1.5mm}
        \caption{Examples from GoEmotions dataset}
        \label{table:1}
        \vspace{-4mm}
        \end{center}
    \end{table}

Humans express emotions via multiple modalities like video (facial expression), speech (intonations and prosody), and text. However, with the growth of the internet, and social media, at a global scale, text is one of the more prominent modes of communication and interaction. With the motivation to perform automatic analysis of emotions expressed in the text, in this paper, we focus on the task of emotion recognition in text. Emotion recognition in text has wide applications \cite{b17}, e.g., marketing \cite{bagozzi1999market}, advertising \cite{Qiu2010adv}, political science \cite{druckman2008polsci}, human-computer interaction \cite{cowie2001hci}, conversational AI \cite{colombo2019affect}.

Formally, given some text, the task of emotion recognition is to predict explicit as well as implicit emotions. Emotions are represented either as a categorical label (e.g., sad, happy, surprise) or as continuous values (in the form of valence and arousal). In this paper, we use the former for representing emotions. Recognizing emotions is challenging as these are often implicit in the text, and the inherent complexity of emotions makes it difficult to model them computationally (see Table \ref{table:1}). 

A number of approaches have been proposed for emotion recognition in text\cite{b1}: keyword-based approaches\cite{tao2004keyword}, rule-based approaches\cite{lee2010rule}, classical learning-based approaches\cite{alm2005classml}, deep learning approaches\cite{meisheri2018deepml}, and hybrid approaches\cite{Gievska2014hybrid}. However, all the proposed approaches follow a two-step method. The first step involves the extraction of emotion-centric features. Subsequently, in the second step, these features are used for identifying emotions via a machine learning-based model. With advances in deep learning, state-of-the-art (SOTA) results have been obtained in almost all major areas, and the task of emotion recognition is not an  exception to this. 

Recently, a new type of neural attention-based deep architecture referred to as Transformer \cite{b11} has been proposed. Transformers excel at handling long-term dependencies in texts. Since the introduction of the initial transformer model, several advanced architectures have been proposed: BERT\cite{b10}, XLNet\cite{b22}, GPT\cite{b23}, RoBERTa\cite{b24}, ALBERT\cite{b25}. Typically, transformers (and their variants) are pre-trained on huge language corpora (e.g., web corpus, Wikipedia), and these pre-trained models can be fine-tuned (adapted) for usage in novel tasks via transfer learning. Pre-training and fine-tuning strategy has shown SOTA results on various NLP tasks like natural language inference, machine comprehension, machine translation, as well as text-based emotion recognition task \cite{b12}. %Transformer based (and its ensembles) models have show SOTA results on text-based emotion recognition task as well \cite{b12}.  

This paper proposes a new transformer-based framework for  fine-grained emotion classification that leverages semantic knowledge of the emotion classes. We use BERT as the base model and attempt to model the semantic meaning of emotion classes through their definitions while training the model for emotion classification. We employ a multi-tasking framework with proportional sampling between emotion classification and definition modeling. We experiment with three setups for definition modeling: 1. Class Definition Prediction (CDP) 2. Masked Language Modeling (MLM) 3. both Class Definition Prediction and Masked Language Modeling(CDP+MLM). 
%Our results
Our model gives an overall improvement in F1 score in all the three setups on the fine-grained GoEmotions dataset \cite{b6}, while the best score is obtained from the setup of CDP. We further test  (via transfer learning) on benchmark datasets with smaller emotion label sets from different domains to check the generalization capability of the proposed model. We use ISEAR\cite{b13}, EmoInt\cite{b14} and Emotion-stimulus\cite{b15}. We observe that our models outperform the existing baselines \cite{b6} in the transfer learning experiments, especially for smaller training data sizes. In summary, we make the following contributions: 
\begin{itemize}
    \item We propose a new framework for emotion prediction that takes into account the definitions of the emotion classes
    \item We obtain the state-of-the-art result for the fine-grained emotion labels prediction task
    \item We perform transfer learning experiments on other datasets from  different domains and label sets and outperform the existing SOTA models for these
\end{itemize}
%Paper outline
We release the code for model implementations and experiments via GitHub: {\url{https://github.com/Exploration-Lab/FineGrained-Emotion-Prediciton-Using-Definitions}}. The following section reviews some of the prominent literature in the area of emotion recognition. Section \ref{sec:methodology} describes the proposed model framework and the training setup. Section \ref{sec:experiments} provides a note on experiments, followed by results in Section \ref{sec:results}. Section \ref{sec:discussion} discusses the observations, improvements, and future work. We conclude the paper in section \ref{sec:conclusion}.
    
    % This document is a model and instructions for \LaTeX.
    % Please observe the conference page limits.
    
\section{Related Work} \label{sec:relatedwork}
    %Emotion prediction
    
Emotion recognition is an actively researched problem, with experts working on it from different fronts in artificial intelligence and cognitive science communities \cite{b16}. %with developments across modalities in the field. 
The text-based challenge in emotion recognition has gained popularity in view of its practical and extensive applications. \cite{b18} evaluated customer satisfaction by comparing the linguistic characteristics of emotional expressions of positive and negative attitudes. \cite{b19} use Twitter data to predict mortality from Atherosclerotic Heart Disease (AHD). There has also been research on the use of emotions to predict personality traits \cite{b20,b21}.

There has been extensive research on representing emotions. These broadly fall into two categories: continuous values and discrete labels. The former is derived from the circumplex model proposed by \cite{b26}; the model states that the space of emotions is continuous, and the emotion experienced can be represented in three dimensions: valence, arousal, and dominance. Based on the circumplex model, various datasets have been created and annotated with three-dimensional continuous values, e.g., EmoBank \cite{b27} and fb-valence-arousal \cite{b28}. There is no single agreed set of discrete emotion labels, giving rise to various emotion label sets. The most common ones are Ekman's six emotions (anger, disgust, fear, joy, sadness, and surprise) \cite{b7} and Plutchik's eight emotions (joy, trust, fear,  surprise,  sadness,  disgust,  anger,  and  anticipation) \cite{b8}. Recent developments have shown that emotions in other modalities may be more fine-grained such as 24 in brief vocalization \cite{b29}, 28 in facial expression \cite{b30}, 12 in speech prosody \cite{b31}. A similar fine-grained emotion label set (27 emotions and neutral) for text has been proposed by \cite{b6}. The authors have created the largest manually annotated dataset (referred to as \textit{GoEmotions}) of emotions in English (details in Section \ref{sec:methodology}). In this work, we propose new models for emotion recognition based on the GoEmotions dataset. \cite{b9} have proposed a \textit{Unified Dataset} for recognizing emotions in text; they combine different emotion datasets into one with a common annotation schema.

%consisting of 58K Reddit %\footnote{\url{www.reddit.com}} 
% comments annotated (via crowd-sourcing) with a set of 27 emotions and neutral. The fine-grained nature of the labels enables capturing subtle changes in emotions, this was not possible in previously proposed datasets. GoEmotions has following 27 emotions: \textit{admiration,  amusement,  anger,  annoyance,approval, caring, confusion, curiosity, desire, disappointment,disapproval,  disgust,  embarrassment,  excitement,  fear,  gratitude, grief, joy, love, nervousness, optimism, pride, realization, relief, remorse, sadness, and surprise.} In this work, we propose new models for emotion recognition on GoEmotions dataset. Due to space constraints, we do not go into details of the annotation scheme and crowdsourcing experiments used for labeling emotions in GoEmotions, we refer the reader to the GoEmotions paper. 

Transformers have dominated recent progress in the NLP community. \cite{b11} introduced attention-based transformer architecture for machine translation which was later adapted to build powerful NLP models independently/jointly using encoder or/and decoder components of the proposed architecture. \cite{b10} presented BERT, which outperformed existing baselines on eleven NLP tasks. The BERT model is based on the encoder component of transformers. It was pretrained on English Wikipedia, and BooksCorpus \cite{b32} using next sentence prediction (NSP) and masked language modeling (MLM) tasks. It was later fine-tuned with additional layers as needed to obtain state-of-the-art results on downstream NLP tasks. In recent years, there have been various attempts in emotion prediction that have used BERT as the base model or in ensemble with some other approach. \cite{b34} investigated the problem of emotion classification in code-switched documents using BERT. The task of emotion classification and constructing emotion lexicon was combined by \cite{b35} that improved emotion classification results on the Twitter dataset. The top three models at the recent EmotionX 2019 \cite{b36} challenge for emotion recognition in text are BERT-based models. The baseline on GoEmotions is also on a BERT model with an added classifier layer fine-tuned on the dataset. 

In the general text classification task using a transformer-based model, an additional layer of a classifier is added to the model, and it is fine-tuned for the task at hand. Here, the semantics of the individual class and its relation with others is not leveraged. There have been some efforts made in this direction lately. \cite{b37} formalized the task of sentence classification as a question answering task where the description of category label and text were concatenated as input for a binary output of yes or no depending on whether the description was correct. This process required the presence of descriptions, and thus, hard attention was used, even during inference. \cite{b38} included four kinds of semantic knowledge (word embeddings, class descriptions, class hierarchy, and a general knowledge graph) in their framework to facilitate zero-shot text classification. \cite{b39} explicitly modeled label embeddings and their mutual correlations in emotion inference for ROC stories \cite{b40}. Related to emotion recognition is the task of Aspect-based sentiment analysis (ABSA) which entails the identification of opinion polarity towards a specific aspect in a given comment. \cite{b41} transformed this into a sentence pair classification task with the concatenation of instance text to an auxiliary sentence constructed with respect to aspect. 

\section{Methodology} \label{sec:methodology}

\noindent\textbf{Dataset:}
%\noindent\textbf{Dataset:} 
In this paper, we use the GoEmotions dataset. Introduced by \cite{b6}, it is the largest manually annotated corpus of 58k English Reddit comments created via crowdsourcing. The text in the corpus is labeled with fine-grained emotion labels consisting of 27 emotions and neutral: \textit{admiration, amusement, anger, annoyance, approval, caring, confusion, curiosity, desire, disappointment, disapproval, disgust, embarrassment, excitement, fear, gratitude, grief, joy, love, nervousness, optimism, pride, realization, relief, remorse, sadness, surprise, neutral}. It is a multi-labeled dataset with 83\% examples marked with one label, 15\% with two labels, 2\% with three labels, and 0.2\% with four or more labels. The dataset is split into train, validation (dev), and test set in $80-10-10$ proportion, respectively. Due to space constraints, we do not go into details of the annotation scheme and crowdsourcing experiments conducted for labeling emotions in GoEmotions; we refer the reader to the GoEmotions paper. 

In our approach, we attempt to model emotion labels' definitions semantics while training the model to predict emotions in a sentence so as to teach the model the descriptive meaning of the emotion classes and their relation to the given instance sentence. We take inspiration from language modeling techniques used to pre-train  BERT. With this goal in mind, we propose a multi-task learning setup, with a primary task of emotion prediction and different auxiliary tasks for learning the definitions. % We model definitions employing tasks inspired from language modelling techniques used to pretrainBERT [4].

\noindent\textbf{Primary Task Definition:} 
%\noindent\textbf{Primary Task Definition:} 
Formally, given a sentence $x$ composed of tokens $\{w_1, w_2, ..., w_n\}$ where n is the length of $x$, the task is to predict the emotion(s) (from a pre-defined set of $L$ discrete emotion labels: $e_{1}, e_{2}, ..., e_{L} $) expressed in the sentence. For example: ``\textit{For art? I think it's kind of silly, but it's also fun. Let people live :P}" would be classified into the category ``joy". Note that a sentence may express more than one emotion, in which case it will be annotated with multiple labels. We are using a transformer-based model for emotion prediction; in accordance with that, the sentence is prepended with a special token $[CLS]$ before passing it to the transformer model. Hence, the sentence becomes: $\{[CLS], w_1, w_2, ..., w_n\}$. Intuitively, the vector representation corresponding to $[CLS]$ token position at the output of the model captures the task-specific information in the sentence. This is further used for emotion prediction. % \cite{b6}.

\begin{figure}[t]
\includegraphics[scale=0.3]{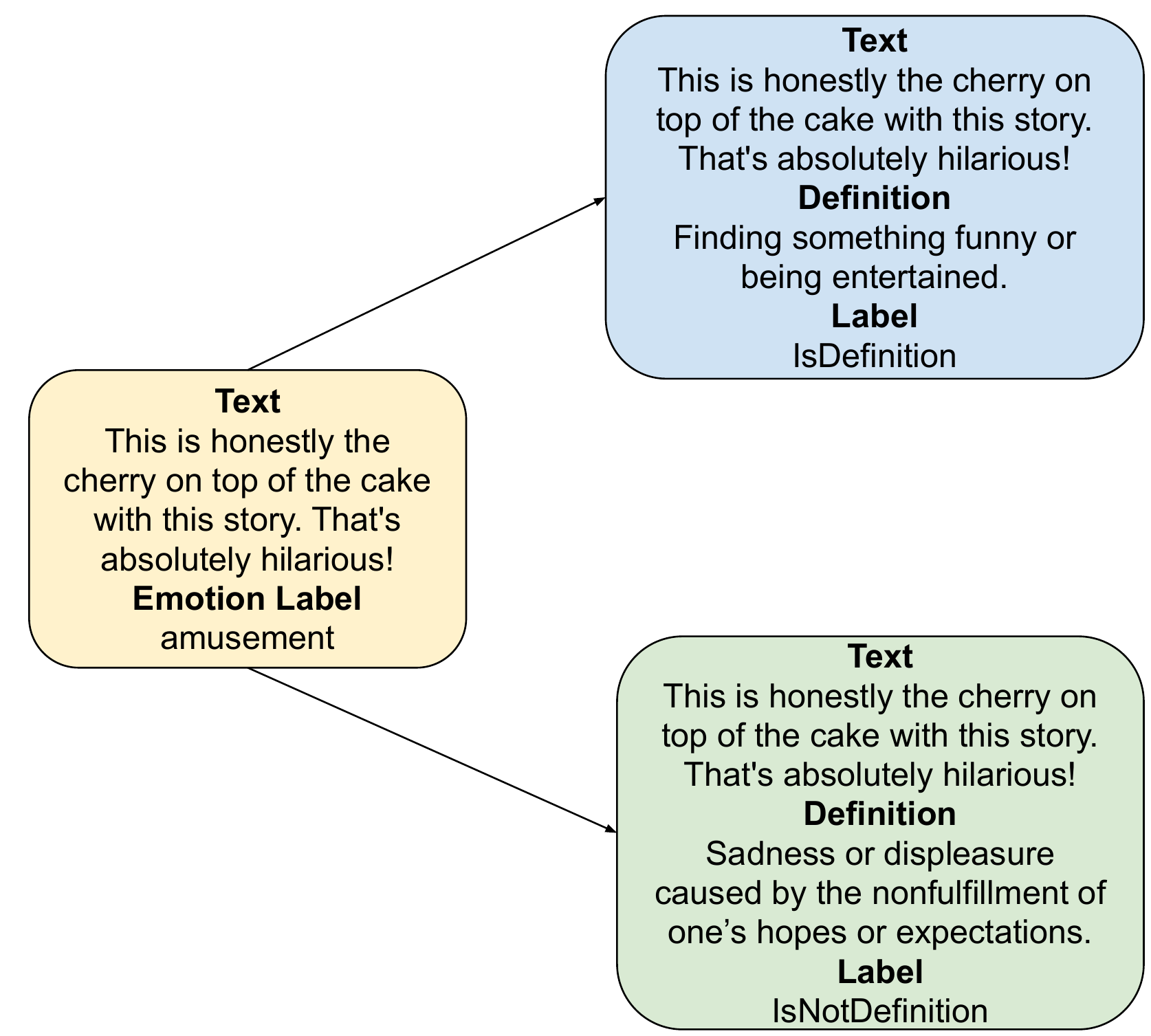}
\centering
\caption{Examples from auxiliary dataset. For each instance in the primary dataset, two instances are constructed for auxiliary dataset; in the first, the text is concatenated with correct emotion's definition and is labeled ``IsDefinition" whereas in the second, the text is concatenated with definition of incorrect emotion label and is labeled ``IsNotDefinition". }
\label{fig:auxins}
\end{figure}

\noindent\textbf{Auxiliary Task Definition - Class Definition Prediction (CDP):}
%\noindent\textbf{Auxiliary Task Definition - Class Definition Prediction (CDP): }
BERT model is trained using the Next Sentence Prediction (NSP) and Masked Language Model (MLM) tasks. In the NSP task, given a pair of sentences, the BERT model predicts if the second sentence semantically follows the first or not. NSP forces the model to learn the semantic correspondences between sentences. Taking inspiration from this, we posit that if a text instance is appended with an emotion label definition and this combination is passed through BERT, then the model would learn the relationship between emotions expressed in the sentence and the corresponding appended emotion label definition. We propose an auxiliary task: given a sequence with a sentence $x$ followed by an emotion definition $c$, the task is to predict if the appended definition depicts the emotions expressed in the text. Formally, we pass the following sequence through BERT model: $\{[CLS], w_{x_{1}}, w_{x_{2}}, ..., w_{x_{n}}, [SEP], w_{c_{1}}, w_{c_{2}}, ..., w_{c_{m}}\}$ where $n$ is the length of the text and $m$ is the length of the definition. Here, $[SEP]$ is a special token that helps the model distinguish between the sentence and the emotion definition. For example, consider the sentence, \textit{``For art? I think it's kind of silly, but it's also fun. Let people live :P"}, we append this with the definition of anger \textit{``A strong feeling of displeasure or antagonism"}. This combination is passed to the BERT model, and in this case, it predicts the ``IsNotDefinition" label (see section \ref{sec:auxdata}),  since the text is not expressing anger.  

% \begin{figure*}
% \centering
% \subfigure{\includegraphics[width=0.3\textwidth]{images/CDP.pdf}} 
% \subfigure{\includegraphics[width=0.3\textwidth]{images/MLM.pdf}} 
% \subfigure{\includegraphics[width=0.3\textwidth]{images/CDP+MLM.pdf}}
% \caption{Model Architecture. In setup 1, only CDP is employed. In setup 2, only MLM is employed. In setup 3, both MLM and CDP are used. Green components correspond to the primary task of emotion prediction only. Red components belong to the auxiliary task only. Yellow components are common between both primary and auxiliary tasks.}
% \label{fig:model}
% \end{figure*}

\begin{figure*}[!h]
\centering
  $\vcenter{\hbox{\includegraphics[height=6.2cm]{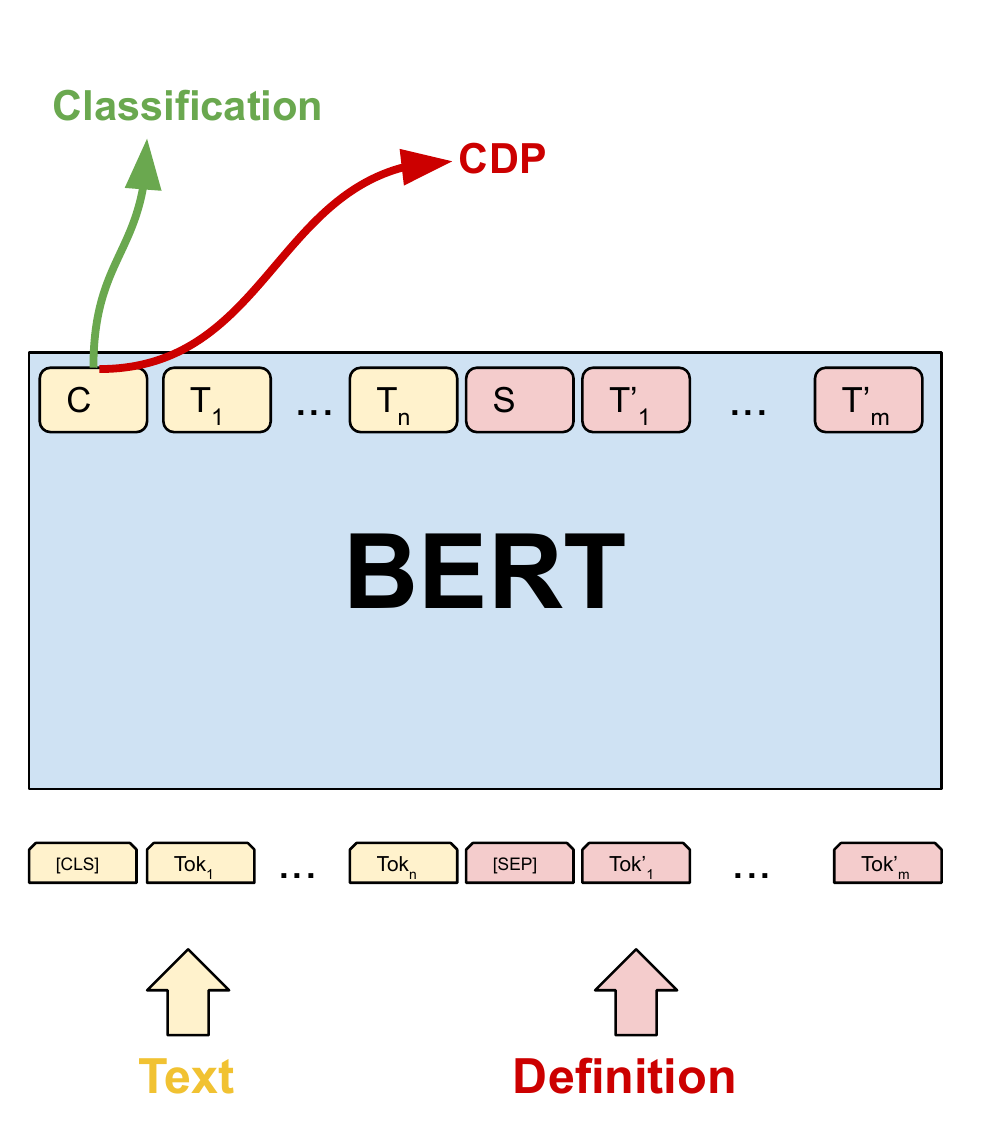}}}$
  \qquad
  $\vcenter{\hbox{\includegraphics[height=6.2cm]{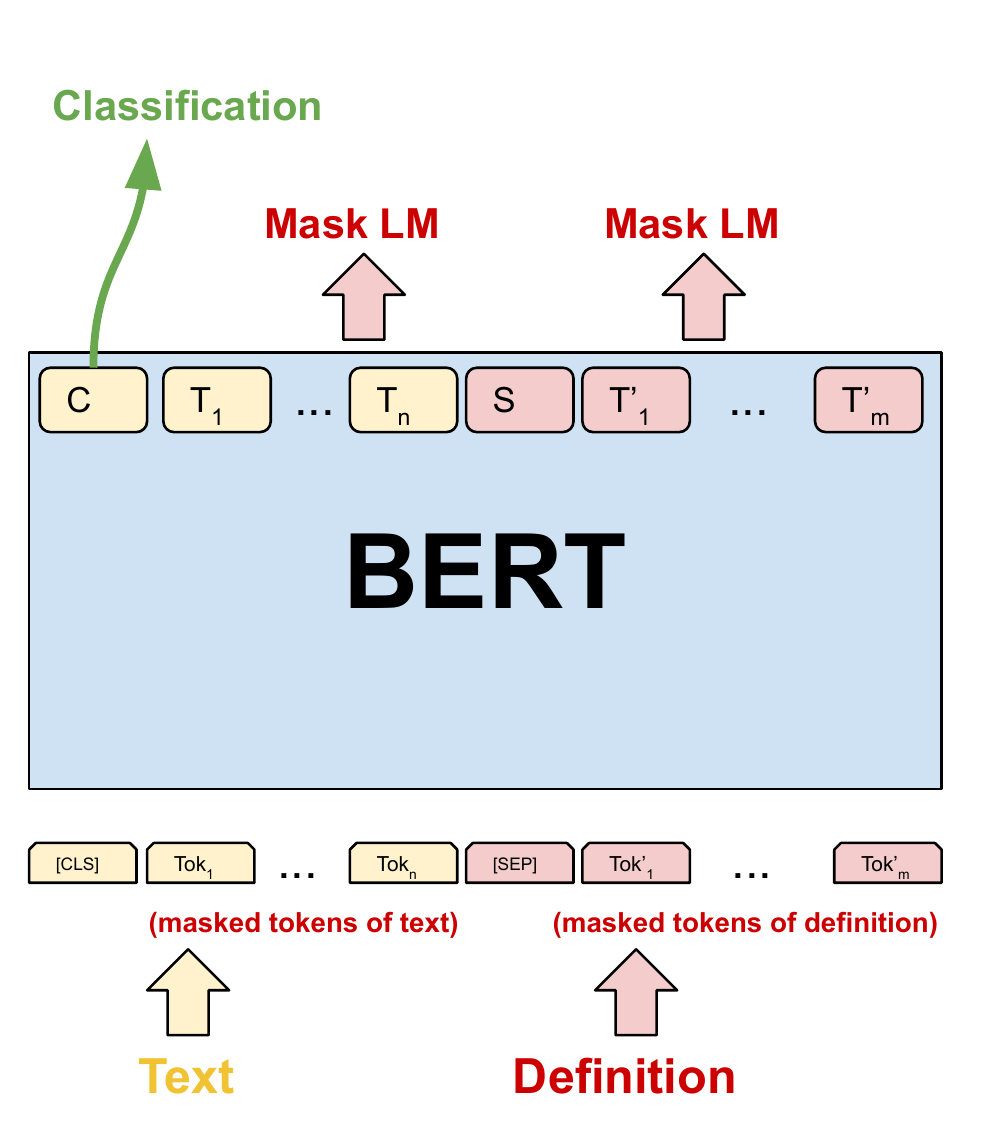}}}$
  \qquad
  $\vcenter{\hbox{\includegraphics[height=6.2cm]{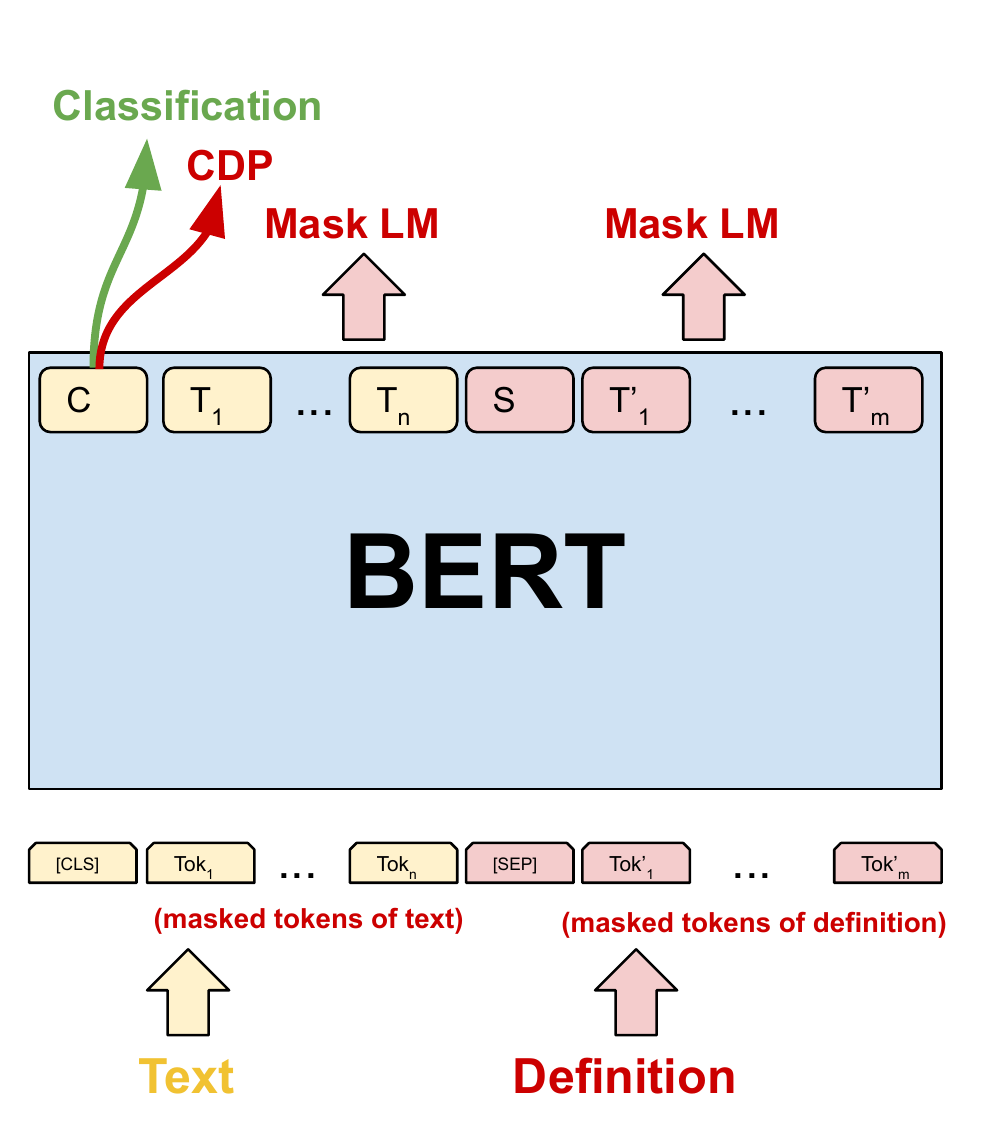}}}$
\caption{Model Architecture. In setup 1, only CDP is employed. In setup 2, only MLM is employed. In setup 3, both MLM and CDP are used. Green components correspond to the primary task of emotion prediction only. Red components belong to the auxiliary task only. Yellow components are common between both primary and auxiliary tasks.}
\label{fig:model}
\end{figure*}

\noindent\textbf{Auxiliary Task Definition - Masked Language Model (MLM):}
%\noindent\textbf{Auxiliary Task Definition - Masked Language Model (MLM): }
In the MLM task, given a sentence, some of the tokens are randomly removed from the sentence and replaced with a special ``$[MASK]$" token. This modified sentence is passed as input to the BERT model, and the task is to predict the $[MASK]$ tokens. This forces the model to understand the semantics of the text to help predict the masked tokens.  We adapt the MLM task to our setting. Given a sequence with a sentence $x$ followed by the definition of one of the emotions depicted in the text $c$ composed of tokens $\{[CLS], w_{x_{1}}, w_{x_{2}}, ..., w_{x_{n}}, [SEP], w_{c_{1}}, w_{c_{2}}, ..., w_{c_{m}}\}$, random tokens are masked by a special token $[MASK]$, where $n$ is the length of the text and $m$ is the length of the definition. The masked tokens are predicted from the vocabulary of the model. For example, given masked input sequence \textit{``[CLS] For art? I think it's kind of silly, but it's also [MASK]. Let people live :P [SEP] A feeling of pleasure and [MASK]"}, correct predictions would be ``fun" and ``happiness" respectively.

\noindent\textbf{Auxiliary Dataset:} \label{sec:auxdata}
%\noindent\textbf{Auxiliary Dataset: } 
For training the auxiliary tasks, we create an auxiliary dataset. The primary dataset, GoEmotions, consists of sentences annotated with emotion labels. For each instance in the dataset, two instances are created in the auxiliary dataset, one labeled `IsDefinition' and the other `IsNotDefinition' (Figure \ref{fig:auxins}). The `IsDefinition' labeled instance contains instance text and definition corresponding to correct emotion as the second sentence in the sequence. The `IsNotDefinition' labeled instance, on the other hand, contains instance text and definition for an emotion not included in the corresponding label as the second sentence. We use the definitions of emotion classes provided by GoEmotions paper authors \cite{b6} to the raters during the annotation process. For multi-labeled instances, each emotion label in the annotation is considered a different instance. 

\noindent\textbf{Model Architecture:} \label{modelarch}
%\noindent\textbf{Model Architecture: } 
We propose a multi-task neural architecture. The primary task is emotion prediction, and the auxiliary task includes CDP or MLM, as described previously. BERT-base model is the backbone for both the primary and auxiliary tasks, i.e., the layers of BERT are shared between the two tasks, primary and auxiliary, with hard parameter sharing (see Figure \ref{fig:model}). The pooled output from BERT is passed through a dropout layer followed by a linear dense classification layer for the primary task. Since it is a multi-label classification problem, sigmoid-based binary cross-entropy is used to compute the loss between the model predictions and class label values. During inference, the text is directly (without the definition) passed to the trained BERT model (primary task), and it is used for prediction directly via the $[CLS]$ token. The probability of each emotion label being present in the given text is predicted by the model. The labels that cross a threshold value (0.3 in our case) are predicted as the emotion for the sentence. Primary and auxiliary tasks are not trained equally; a proportional sampling method is employed to sample between the two tasks. During each iteration, we sample from a Bernoulli distribution: $Bernoulli(p)$,  to decide between primary and auxiliary tasks to train. Here, the  probability $p$ corresponds to the selection of the primary task. We experiment with three different auxiliary tasks (Figure \ref{fig:model}):
\begin{enumerate}[noitemsep,topsep=0pt]
\item \textbf{CDP: } The pooled output from the BERT-base model is input to a dense linear classification layer followed by computation of cross-entropy loss. 
\item \textbf{MLM: } Only the positive (`IsDefinition' labeled) instances of the auxiliary dataset are used for this setup. The input sentence-pair tokens are masked by following the same strategy used by BERT. A softmax activation function is used over the tokens in vocabulary for predicting the masked tokens using the hidden vectors in output from BERT. This is also followed by cross-entropy loss.
\item \textbf{CDP+MLM: } Model is trained on both CDP and MLM tasks in this setup, and all the instances of the auxiliary dataset are included in the training. A combined loss from both CDP and MLM is used.
\end{enumerate}

\section{Experiments} \label{sec:experiments}
\noindent\textbf{Different Transformer Architectures Experiments:} 
We experiment with different transformer architectures: BERT, XLNet, RoBERTa, ALBERT, to compare their performances as base models for emotion classification on GoEmotions dataset. We keep the same hyperparameters as in \cite{b6}, except for an increase in the number of training epochs to 10. A classifier layer is added to all the models, followed by binary cross-entropy loss. Binary classification on all categories is used to accommodate multi-labeled instances.

\noindent\textbf{Fine-Grained Experiments:} 
We use the same hyperparameters as in the previous experiment. We experiment with proportional sampling during training with the following range of values for Bernoulli distribution probability $p \in \{0.1, 0.2, 0.3, ..., 0.9\}$, where $p$ corresponds to the probability of training the primary task during an iteration. %The training task is randomly chosen with respect to this probability distribution. 
These experiments are carried on all the three auxiliary tasks setups considered for the proposed framework as explained in Section \ref{modelarch}:
\begin{enumerate}
    \item CDP only
    \item MLM only
    \item both CDP and MLM
\end{enumerate}

\noindent\textbf{Transfer Learning-Based Experiments:} 
In transfer learning-based experiments, we test the adaptability and generalization capability of the proposed model to other domains and emotion label sets. We use the best performing model(s) trained on GoEmotions dataset obtained in previous experiments to initialize training on other datasets, thereby \textit{transferring} the knowledge learned from GoEmotions to a target domain. We compare the performance of our model(s) to vanilla BERT and BERT trained on GoEmotions proposed by \cite{b6} following the same parameters and experimental setup. Emotion corpora from the \textit{Unified Dataset} \cite{b9} are selected for the experiments; in particular, we experiment on benchmark datasets ISEAR\cite{b13}, EmoInt\cite{b14}, and Emotion-Stimulus\cite{b15} owing to their diversity in source domains and emotion label sets. During training on each dataset, train data size is varied in the range $(100, 200, 500, 1000, 80\%\ of\ total)$, the remaining data is taken for the test set. The above is repeated for ten random splits for each set size. 

\begin{table*}[]
\begin{center}
\begin{tabular}{lcccccc}
\hline
\textbf{Model} & \multicolumn{1}{l}{\textbf{Dev-Precision}} & \multicolumn{1}{l}{\textbf{Dev-Recall}} & \multicolumn{1}{l}{\textbf{Dev-F1}} & \multicolumn{1}{l}{\textbf{Test-Precision }} & \multicolumn{1}{l}{\textbf{Test-Recall }} & \multicolumn{1}{l}{\textbf{Test-F1}} \\ \hline
Baseline       & -                                          & -                                       & -                                   & 40                                          & 63                                       & 46                                   \\ \hline
BERT           & 53.84                                      & 50.13                                   & 50.28                               & 48.86                                       & 53.11                                    & 50.17                                \\
XLNet          & 49.3                                       & 52.89                                   & 50.27                               & 47.09                                       & 51.92                                    & 48.5                                 \\
RoBERTa        & 51.82                                      & 53.76                                   & 50.87                               & 51.82                                       & 52.94                                    & 51.21                                \\
ALBERT         & 54.56                                      & 50.12                                   & 50.03                               & 56.98                                       & 48.89                                    & 50.12                                \\ \hline
BERT+CDP       & 53.57                                      & 53.21                                   & 52.07                               & 54.66                                       & 53.8                                     & 52.34                                \\
BERT+MLM       & 52.04                                      & 53.13                                   & 51.57                               & 52.42                                       & 52.67                                    & 51.25                                \\
BERT+CDP+MLM   & 53.29                                      & 54.12                                   & 52.02                               & 53.42                                       & 54.37                                    & 51.96                                \\ \hline

\end{tabular}
\vspace*{1mm}
\caption{Performance (F1-Score (in \%)) of baseline, transformers and proposed models on dev and test split of GoEmotions}
\label{restable}
\end{center}
\end{table*}

\begin{table*}[]
\begin{center}
\begin{tabular}{ccccccc}
\hline
\textbf{Sampling Probability ($p$)} & \multicolumn{2}{c}{\textbf{BERT+CDP}}                                & \multicolumn{2}{c}{\textbf{BERT+MLM}}                                & \multicolumn{2}{c}{\textbf{BERT+CDP+MLM}}                            \\ \hline
                            & \multicolumn{1}{c}{\textbf{Dev}} & \multicolumn{1}{c}{\textbf{Test}} & \multicolumn{1}{c}{\textbf{Dev}} & \multicolumn{1}{c}{\textbf{Test}} & \multicolumn{1}{c}{\textbf{Dev}} & \multicolumn{1}{c}{\textbf{Test}} \\ \hline
0.1              & 44.57                            & 44.05                             & 43.1                             & 42.74                             & 45.33                            & 44.98                             \\
0.2               & 47.24                            & 46.18                             & 48.35                            & 48.19                             & 48.71                            & 48.3                              \\
0.3               & 47.61                            & 47.38                             & 49.73                            & 49.11                             & 51.49                            & 50.36                             \\
0.4              & 51.65                            & 51.07                             & 51.1                             & 50.07                             & 51.33                            & 50.87                             \\
\textbf{0.5}              & \textbf{52.07}                            & \textbf{52.34}                             & \textbf{51.57}                            & \textbf{51.25}                             & \textbf{52.02}                            & \textbf{51.96}                             \\
0.6               & 50.98                            & 52.32                             & 49.92                            & 50.27                             & 51.32                            & 49.72                             \\
0.7               & 52.13                            & 51.2                              & 49.57                            & 50.6                              & 50.76                            & 50.09                             \\
0.8               & 51.32                            & 51.73                             & 49.5                             & 49.51                             & 49.88                            & 49.43                             \\
0.9               & 50.35                            & 50.79                             & 50.34                            & 49.77                             & 49.85                            & 51.26                             \\ \hline

\end{tabular}
\vspace*{1mm}
\end{center}
\caption{Comparison of F1 scores (in \%) of different proportion samplings for BERT+CDP, BERT+MLM, BERT+CDP+MLM on dev and test split of GoEmotions}
\label{tab:prop-res}
\end{table*}

\begin{table*}[]
\begin{center}
\begin{tabular}{lccccc}
\hline
\textbf{Emotion}       & \textbf{Baseline} & \textbf{BERT} & \textbf{BERT+CDP} & \textbf{BERT+MLM} & \textbf{BERT+CDP+MLM} \\ \hline
admiration             & 65                           & 66.85                    & 66.54                        & 67.42                        & 68.56                            \\
amusement              & 80                           & 80.27                    & 81.26                        & 82.47                        & 81.8                             \\
anger                  & 47                           & 49.64                    & 48.93                        & 49.65                        & 50                               \\
annoyance              & 34                           & 33.43                    & 36.43                        & 36.26                        & 39.51                            \\
approval               & 36                           & 35.63                    & 39.8                         & 41.75                        & 39.12                            \\
caring                 & 39                           & 43.48                    & 41.09                        & 43.26                        & 40.92                            \\
confusion              & 37                           & 44.32                    & 39.48                        & 46.88                        & 41.62                            \\
curiosity              & 54                           & 51.77                    & 55.84                        & 56.24                        & 55.62                            \\
desire                 & 49                           & 47.8                     & 48.89                        & 52.29                        & 47.8                             \\
disappointment         & 28                           & 27.91                    & 30.34                        & 33.33                        & 30.09                            \\
disapproval            & 39                           & 41.26                    & 39.66                        & 41.37                        & 36.49                            \\
disgust                & 45                           & 43.82                    & 50.86                        & 48                           & 46.15                            \\
\textbf{embarrassment}          & \textbf{43}                           & \textbf{43.59}                    & \textbf{50.85}                        & \textbf{50}                           & \textbf{50.79}                            \\
excitement             & 34                           & 43.95                    & 39.2                         & 41.95                        & 41.15                            \\
fear                   & 60                           & 65.9                     & 65.91                        & 62.5                         & 64.68                            \\
gratitude              & 86                           & 89.79                    & 91.47                        & 91.49                        & 91.73                            \\
\textbf{grief}                  & \textbf{0}                            & \textbf{25}                       & \textbf{50}                           & \textbf{0}                            & \textbf{50}                               \\
joy                    & 51                           & 60.57                    & 57.8                         & 61.86                        & 58.22                            \\
love                   & 78                           & 79.09                    & 77.69                        & 79.84                        & 78.49                            \\
\textbf{nervousness}            & \textbf{35}                           & \textbf{37.5}                     & \textbf{43.9}                         & \textbf{45.45}                        & \textbf{44.44}                            \\
optimism               & 51                           & 55.47                    & 55.61                        & 54.84                        & 56.58                            \\
\textbf{pride}                  & \textbf{36}                           & \textbf{41.67}                    & \textbf{34.78}                        & \textbf{45.45}                        & \textbf{43.48}                            \\
realization            & 21                           & 21.25                    & 25                           & 24.1                         & 25.58                            \\
\textbf{relief}                 & \textbf{15}                           & \textbf{33.33}                    & \textbf{50}                           & \textbf{40}                           & \textbf{40}                               \\
remorse                & 66                           & 64.12                    & 67.18                        & 67.14                        & 64.75                            \\
sadness                & 49                           & 55.93                    & 56.86                        & 53.09                        & 47.94                            \\
surprise               & 50                           & 54.98                    & 54.05                        & 53.1                         & 53.92                            \\
neutral                & 68                           & 66.49                    & 66.17                        & 65.33                        & 65.4                             \\ \hline
\textbf{macro-average} & 46                           & 50.17                    & 52.34                        & 51.25                        & 51.96                            \\ \hline
\textbf{std}           & 19                           & 16.95                    & 15.6                         & 18.14                        & 15.47                            \\ \hline

\end{tabular}
\vspace*{1mm}
\end{center}
\caption{Comparison of F1 scores (in \%) of each emotion category from the  GoEmotions dataset. The highlighted categories have less training examples in the GoEmotions dataset, however the proposed model shows performance improvements.}
\label{tab:indv-emo-res}
\end{table*}

\begin{figure*}[!h]
\centering
  $\vcenter{\hbox{\includegraphics[height=4.4cm]{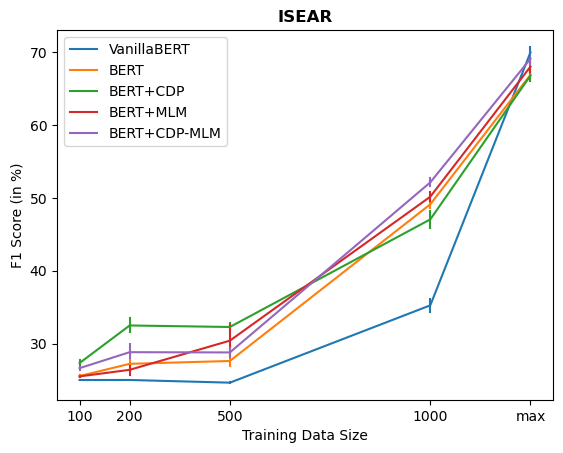}}}$
  \qquad
  $\vcenter{\hbox{\includegraphics[height=4.4cm]{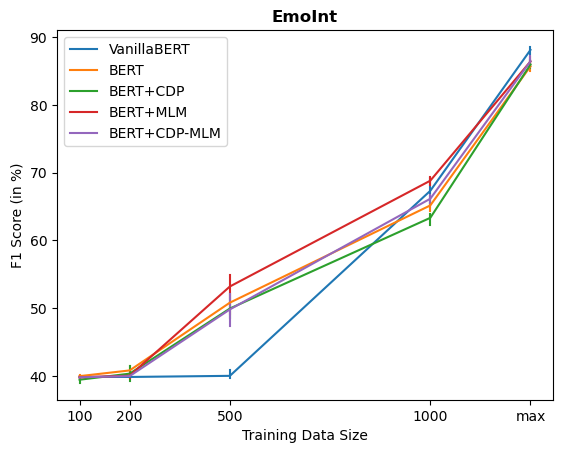}}}$
  \qquad
  $\vcenter{\hbox{\includegraphics[height=4.4cm]{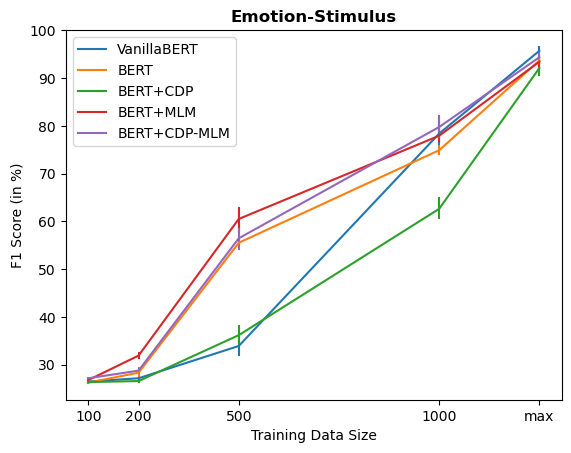}}}$
    \caption{Transfer learning experiments on ISEAR, EmoInt and Emotion-Stimulus datasets with differing initialisation of base model BERT for finetuning. VanillaBERT is pretrained BERT, BERT refers to pretrained BERT finetuned on GoEmotions. The other three variants (BERT+CDP, BERT+MLM, BERT+CDP+MLM) refer to the different frameworks finetuned on GoEmotions using definitions.}
    \label{fig:transferplots}
\end{figure*}

% \begin{figure*}
%     \centering
%     \subfigure{\includegraphics[width=0.3\textwidth]{images/isear_output_f1_final.png}} 
%     \subfigure{\includegraphics[width=0.3\textwidth]{images/emoint_output_f1_final.png}} 
%     \subfigure{\includegraphics[width=0.3\textwidth]{images/emotion-cause_output_f1_final.png}}
%     \caption{Transfer learning experiments on ISEAR, EmoInt and Emotion-Stimulus datasets with differing initialisation of base model BERT for finetuning. VanillaBERT is pretrained BERT, BERT refers to pretrained BERT finetuned on GoEmotions. The other three variants (BERT+CDP, BERT+MLM, BERT+CDP+MLM) refer to the different frameworks finetuned on GoEmotions using definitions.}
%     \label{fig:transferplots}
% \end{figure*}

\section{Results} \label{sec:results}

Standard Precision, Recall, and F1-Score metrics are used for evaluating various models.

\noindent\textbf{Different Transformer Architectures:} 
The results on various transformer architectures are shown in Table \ref{restable}. The baseline model is a BERT model as proposed for GoEmotions dataset \cite{b6}. All the transformer models are trained for 10 epochs. However, the baseline model was trained for only 4 epochs by \cite{b6}. As evident from the results, training for more epochs improves the model performance. We also observe that the F1-scores for RoBERTa and ALBERT are similar to BERT, whereas the XLNet model performs slightly worse.

\noindent\textbf{Fine-Grained Experiments:} We experimented with various proportional sampling probabilities. The results for different probabilities are shown in \ref{tab:prop-res}.  We found that a proportion sampling probability of $0.5$ for primary tasks works best in all three setups of the framework. The best-performing models of each setup outperform the existing baseline, and BERT+CDP provides the best results overall. Table \ref{tab:indv-emo-res} provides the performance of the best model on each of the emotion categories. We find that our BERT+CDP+MLM model framework improves the performance significantly among most categories. In particular, the improvement for the following categories (with fewer training data examples $( < 500 )$) are significant: \textit{embarrassment, grief, nervousness, pride, relief}.

\noindent\textbf{Transfer Learning Experiments:} 
Figure \ref{fig:transferplots} shows the plot for results on different datasets for various train dataset sizes. We find that GoEmotions-trained BERT+MLM initialized models consistently outperform the existing baseline of GoEmotions-trained BERT on all the datasets. For smaller dataset sizes, models employing transfer learning from GoEmotions give better results as compared to just pretrained BERT, but the difference decreases with increment in training data size. 

\section{Discussion and Error Analysis} \label{sec:discussion}

%task, contributions and implications
BERT+CDP gives the best results outperforming the earlier baselines in \cite{b6}, but BERT+CDP+MLM also has a competitive performance; BERT+MLM does not perform as well but still performs better than the baseline BERT model. In the results for individual categories, we notice an increment in the scores for categories with fewer examples (\textit{embarrassment, grief, nervousness, pride, relief}) in training for BERT+CDP+MLM setup, which shows the relevance of our model with unbalanced training data. A majority of the other emotion categories' scores also improve. We observe that the GoEmotions trained BERT+MLM transferred models perform consistently better than the baseline in the transfer learning experiment. There are variations across the three datasets, which might be due to their diverse nature in terms of the source domain, emotion labels, and balance across classes for training examples. 

Emotion classification is one of the more challenging tasks in text classification not just because of the abstract nature of emotions but also because of the possible subjectivity in interpretation. Unlike a number of NLP tasks, problems in the affective domain cannot always be determined through linguistic cues alone. We notice this issue in our error analysis, where we find that some predictions of the trained model, though incorrect as per annotations, could be reasonable from a different perspective. We present examples of some of these instances in Table \ref{insttable}. These instances, among many others, though judged as erroneous predictions, make sense. This calls for developing methods for addressing the subjectivity of emotion labels and annotation uncertainty due to it \cite{zhang2018dynamic, rizos2019modelling}; we plan to explore this in the future work.  
\begin{table}[H]
\begin{tabular}{lll}
\hline
\textbf{Text}  & \textbf{Label(s)} & \textbf{Prediction(s)} \\ \hline
\begin{tabular}[c]{@{}l@{}}I’m really sorry about your situation :(\\  Although I love the names Sapphira,\\  Cirilla, and Scarlett!\end{tabular} & sadness           & \begin{tabular}[c]{@{}l@{}}love, \\ remorse\end{tabular}             \\ \hline
\begin{tabular}[c]{@{}l@{}}Kings fan here, good luck to you guys!\\  Will be an interesting game to watch!\end{tabular}                           & excitement        & \begin{tabular}[c]{@{}l@{}}excitement, \\ optimism\end{tabular}      \\ \hline
Boomers ruined the world                                                                                                                          & neutral           & \begin{tabular}[c]{@{}l@{}}annoyance, \\ disappointment\end{tabular} \\ \hline
\begin{tabular}[c]{@{}l@{}}Now I'm wondering if {[}NAME{]} drinks,\\ and if he's ever been inebriated during\\  one of his deals.\end{tabular}   & surprise          & \begin{tabular}[c]{@{}l@{}}curiosity, \\ surprise\end{tabular}       \\ \hline
\begin{tabular}[c]{@{}l@{}}I totally thought the same thing! I was\\  like, oh honey nooooo!\end{tabular}                                         & neutral           & realization                                                          \\ \hline
\end{tabular}
\vspace*{1mm}
\caption{Examples of mismatch between the predictions by the model and average predictions by annotators. This is primarily due to the subjective nature of emotions.}
\label{insttable}
\end{table}

%observations and future dev
\section{Conclusion} \label{sec:conclusion}
In this paper, we addressed the task of fine-grained emotion prediction. The key idea is to use emotion label definitions for prediction. We employed language modeling tasks inspired from work by \cite{b10} to model definitions of emotion classes as an auxiliary task to inform the model of the affective meaning of the emotion conveyed in a sentence. We tested our model with the two auxiliary tasks: class definition prediction (CDP) and masked language modeling (MLM), in both isolation and combined format, thereby giving us three different architectures and experimental setups with BERT transformer as the base model and primary task emotion classification. We obtain the state-of-the-art result for fine-grained labels of the GoEmotions dataset consisting of 27 emotions and neutral. Lastly, the proposed state-of-the-art model also \textit{transfers} knowledge well and outperforms the existing baseline of transfer learning to other datasets. Fine-grained emotions can be represented as taxonomy, and this could be used for learning the relationships between different emotion categories. This could, in turn, be useful for the general task of emotion recognition. In the future, we plan to pursue this line of research. %In the future, we plan to extend our framework to other tasks in text classification. 

\bibliographystyle{IEEEtran}
\bibliography{references}
\end{document}